# When Visual Privacy Protection Meets Multimodal Large Language Models

Xiaofei Hui[1] · Qian Wu[2] · Haoxuan Qu[1] · Majid Mirmehdi[3] · Hossein Rahmani[1] · Jun Liu[1]

**Abstract**
The emergence of Multimodal Large Language Models (MLLMs) and the widespread usage of MLLM cloud services such as GPT-4V raised great concerns about privacy leakage in visual data. As these models are typically deployed in cloud services, users are required to submit their images and videos, posing serious privacy risks. However, how to tackle such privacy concerns is an under-explored problem. Thus, in this paper, we aim to conduct a new investigation to protect visual privacy when enjoying the convenience brought by MLLM services. We address the practical case where the MLLM is a "black box", i.e., we only have access to its input and output without knowing its internal model information. To tackle such a challenging yet demanding problem, we propose a novel framework, in which we carefully design the learning objective with Pareto optimality to seek a better trade-off between visual privacy and MLLM's performance, and propose critical-history enhanced optimization to effectively optimize the framework with the black-box MLLM. Our experiments show that our method is effective on different benchmarks.

**Keywords** VQA · MLLM · Action Recognition · Video Analysis

## 1 Introduction

Recently, the emergence and advancement of Multimodal Large Language Models (MLLMs) such as GPT-4V Achiam

Communicated by Boxin Shi.

✉ Jun Liu
j.liu81@lancaster.ac.uk

Xiaofei Hui
x.hui@lancaster.ac.uk

Qian Wu
fivethousand@tongji.edu.cn

Haoxuan Qu
h.qu5@lancaster.ac.uk

Majid Mirmehdi
m.mirmehdi@bristol.ac.uk

Hossein Rahmani
h.rahmani@lancaster.ac.uk

[1] School of Computing and Communications, Lancaster University, South Dr,, Lancaster LA1 4WA, United Kingdom

[2] Singapore University of Technology and Design, 8 Somapah Rd, Singapore 487372, Singapore

[3] Department of Computer Science, University of Bristol, Queens Road, Bristol BS8 1QU, United Kingdom

et al. (2023) and Liu et al. (2023a) has notably improved the convenience of various daily tasks. These MLLMs combine the reasoning power of text instructions with the interpretive capabilities of images and videos, and have shown their superiority in various tasks such as scene understanding and human activity analysis (Liu et al., 2023b; Li et al., 2023). Due to their convenient services and comprehensive capabilities, it has become a popular trend to use MLLMs as helpers to solve various problems in different aspects of our daily life Yin et al. (2023).

The widespread adoption of MLLMs as cloud-based services has raised growing concerns about user privacy (Wu et al., 2023). While uploading images and videos enables convenient access to these services, it may inadvertently expose sensitive personal information that users did not intend to share, such as facial identity, skin tone, gender attributes, and location cues (Orekondy et al., 2017; Wu et al., 2020). Moreover, user-uploaded data may be further collected and reused for model training and refinement (OpenAI, 2026; Gemini Apps Privacy Hub, 2026), potentially amplifying privacy risks (Nasr et al., 2023). In response to these concerns and potential violations of privacy regulations (Justic & Liberties, 1974; Eu) & , 2016), several countries have initiated investigations and imposed restrictions on the deployment

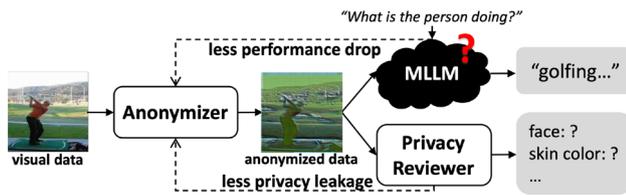

**Fig. 1** Illustration of our problem. When using cloud MLLM services, directly uploading visual data poses privacy risks. To tackle this problem, we propose to train an anonymizer to remove the privacy information in the visual data (as assessed by a privacy reviewer) while maintaining black-box MLLM's performance. This shall allow users to preserve privacy while still benefiting from the MLLM services

of MLLM services (ChatGPT banned in Italy over privacy concerns., 2023; OPC to investigate ChatGPT jointly with provincial privacy authorities, 2023).

Urgent attention is needed to address the critical issue of visual privacy when using MLLM services. However, current privacy-preservation methods typically fall short in achieving a robust privacy-utility balance in black-box MLLM scenarios. Specifically, some privacy preservation approaches protect privacy by degrading image and video quality (Butler et al., 2015; Dai et al., 2015) or by modifying the regions containing privacy information (Ren et al., 2018; Zhang et al., 2021). While such methods can reduce privacy-leak risks, they often considerably damage the information in the visual data, and this can degrade the performance of the utility model (e.g. action recognition model) Dave et al. (2022); Peng et al. (2023). Some other methods have aimed at specifically preserving the utility model's performance while protecting visual privacy (Wu et al., 2018; Li et al., 2023; Fioresi et al., 2023). Generally, they train an anonymization model that learns to anonymize the visual data, and at the same time, a utility model that learns to perform the task with the anonymized data. Although such methods can achieve a good trade-off between maintaining privacy and task performance, they generally require training and modification of the utility model, necessitating full access to its structure and parameters. This is hardly feasible for an MLLM as the utility model, since (i) many cutting-edge MLLM cloud services, such as GPT-4V, are not open access, and (ii) even with access to model parameters, training multi-billion parameter MLLM models is extremely computationally expensive. As a result, there is an urgent need for novel and effective solutions that address privacy concerns in the context of emerging MLLMs.

A practical approach to this problem is to treat the MLLM as a "black-box" function, i.e., we cannot access its internal structure and parameters, only its inputs and outputs, and can only perform forward (inference) processes with the frozen MLLM. In such a scenario, as shown in Fig. 1, we propose to train an anonymizer to protect visual privacy such that the privacy information (e.g., face and skin color) is maximally removed, while the performance of the MLLM is least affected. However, training such an anonymizer is challenging, as the anonymized visual data must remain *directly understandable* by an off-the-shelf MLLM trained solely on raw visual data, while no gradient information can be obtained from the black-box MLLM to guide learning.

To handle the above challenges, in this paper, we propose a novel privacy preservation framework for black-box MLLMs. In this framework, we explore how to effectively achieve a balanced trade-off between privacy and utility of the black-box MLLM. Naively, to achieve this, we can adopt zeroth-order optimization (Spall, 1992) to estimate gradients for the (black-box) utility objective, and train the anonymizer with a simple weighted sum of the privacy and utility objectives. Yet, the estimated gradients for the black-box utility objective have distinct characteristics compared to the computed gradients for the privacy objective, and simply using the weighted sum with these unbalanced gradients leads to inferior performances (see Tab. 5). Thus, we analyze this problem using Pareto optimality, and propose a tailored design for the overall objective that achieves a balanced trade-off between privacy and black-box MLLM's utility.

Moreover, in this process, we also aim to handle other optimization challenges posed by the black-box MLLM. Specifically, zeroth-order optimization such as SPSA Spall (1992) can estimate gradients, yet, it typically involves randomness in the optimization, causing unstability and requires numerous trials to finally reach convergence (as shown in Fig. 3). This does not fit well to our problem, as each of the trials requires expensive cloud MLLM API calls, leading to large training costs. Thus, we propose a critical-history enhanced optimization scheme that reduces the random errors by aggregating critical history gradients along the optimization path and using them as prior information to facilitate the learning process. This scheme, though simple, can quite effectively stabilize and accelerate the convergence, improving the performance with less MLLM queries needed in the training process.

Our major contributions can be summarized as follows: (1) We propose a novel privacy preservation framework that trains an anonymizer directly with black-box MLLMs without requiring access to model parameters or gradients, addressing the critical problem of protecting user privacy in visual data when using MLLM services. (2) In this framework, we propose to frame the privacy-utility trade-off as a multi-objective optimization problem and introduce an augmented Tchebycheff objective to seek Pareto-optimal solutions, providing a principled way to balance privacy protection and MLLM utility. (3) Moreover, considering the emergent optimization challenges in the presence of black-box MLLM, we propose a critical-history enhanced optimization scheme that enables efficient and stable training of the anonymizer with expensive black-box MLLM queries. We evaluate our proposed framework on privacy-preservation tasks using popular MLLMs. Our results demonstrate its effectiveness with different utilization tasks.

## 2 Related Work

**Privacy Preservation** methods can be roughly categorized into the following: (1) Downsampling methods (Butler et al., 2015; Dai et al., 2015; Liu & Zhang, 2020) intentionally reduce the quality of images and videos to conceal privacy information. Though straightforward, it can be hard for these methods to balance the privacy-utility trade-off (Li et al., 2023). (2) Obfuscation-based methods (Ren et al., 2018; Zhang et al., 2021; Ilic et al., 2024) generally first localize the privacy regions with an off-the-shelf model, and then modify the localized content to remove privacy information. These methods depend crucially on the model being able to correctly localize the privacy regions in the first step, limiting their performance. (3) Recently, adversarial training methods (Wu et al., 2018, 2020; Li et al., 2023; Peng et al., 2023; Dave et al., 2022; Fioresi et al., 2023; Kumawat & Nagahara, 2022) are studied to find a better trade-off between privacy and utility. These methods generally employ an anonymization model to generate anonymized data, a privacy attribute classification model to examine privacy leakage in the anonymized data, and a utility model specifically trained to perform the task with the anonymized data. Though these methods achieve a better trade-off, they typically require full access to the utility model to specifically adapt it to perform tasks with anonymized data, which thus is not suitable for handling the black-box utility model in our problem. Differently, we are the first to investigate protecting visual privacy when using black-box MLLM services, and propose a novel privacy-preserving framework with black-box MLLM to achieve a better trade-off between privacy and utility.

**Multimodal Large Language Models (MLLMs)** have witnessed much progress recently (Achiam et al., 2023; Liu et al., 2023a; Liang et al., 2024), with a trend to build and deploy MLLMs for real-world applications, such as video interpretation (Lin et al., 2023; Maaz et al., 2023), robotic planning (Lv et al., 2024) and healthcare diagnostic (Liu et al., 2023). However, as there can be a large amount of visual privacy information in such applications, there have also been widespread privacy concerns (ChatGPT banned in Italy over privacy concerns., 2023; OPC to investigate ChatGPT jointly with provincial privacy authorities, 2023). In this paper, we study protecting privacy information in visual data when using MLLM services.

**Zeroth-order Optimization** has been explored in areas such as control systems (Lu et al., 2015; Li et al., 2023), image classification (Oh et al., 2023), and signal processing (Liu et al., 2020). Different from these works, we consider the complex MLLM as a black box and tackle the challenging problem of preserving its visual privacy. Optimization with black-box MLLM can be extremely difficult in our problem, as zeroth-order optimization typically involves estimation errors, especially when optimizing over such a complex system as MLLM. This unfortunately leads to slow and unstable optimization, and consequently requires numerous API calls of the MLLM cloud service in our problem, which can be costly. Thus, we specifically design a novel optimization scheme to handle the challenges in zeroth-order optimization with black-box MLLM.

## 3 Method

### 3.1 Preliminaries

**Pareto Optimality.** We first introduce the definitions for Pareto dominance and Pareto optimality (Miettinen, 1999; Steuer & Choo, 1983) below:

**Definition 1** *(Pareto Dominance)* Consider $n$ objectives, and let $\theta_1, \theta_2 \in \mathbb{R}^{p_A}$. $\theta_1$ is said to dominate $\theta_2$, if and only if $L_i(\theta_1) \leq L_i(\theta_2)$, $\forall i \in \{1, ..., n\}$, and besides $L_j(\theta_1) < L_j(\theta_2)$, $\exists j \in \{1, ..., n\}$.

**Definition 2** *(Pareto Optimality)* A solution $\theta^* \in \mathbb{R}^{p_A}$ is Pareto optimal, if there exists no $\theta$ that dominates $\theta^*$.

In other words, if a solution is Pareto optimal, it indicates that there are no other solutions that can improve one objective without hurting the other one, i.e., this solution is an optimal trade-off between the two objectives.

**Tchebecheff norm.** The Tchebycheff norm is a scalarization technique in that minimizes the maximum weighted deviation from an ideal point across objectives:

$$L_{tche}(\theta) = \max_i \{\lambda_i (L_i(\theta) - (z_i^* - \varepsilon))\}, \qquad (1)$$

where $\sum_i \lambda_i = 1$, $i \in \{1, 2\}$, $z_i^*$ is the ideal value of the objective, and $(z_i^* - \varepsilon)$ is designed to be the unachievable ideal point with $\varepsilon$ being a small positive scalar to ensure numerical stability and prevents the optimization from getting stuck when the values of the objectives approach the ideal values. Here, $(L_i(\theta) - (z_i^* - \varepsilon))$ measures the distance between the current objective and the ideal value, and Eq. 1 essentially returns the objective that is more distant from its ideal point at the current step.

**SPSA.** Simultaneous Perturbation Stochastic Approximation (SPSA) (Spall, 1992) is a gradient-free optimization method for black-box optimization scenarios where gradients are unavailable. Specifically, to optimize parameters $\theta \in \mathbb{R}^p$

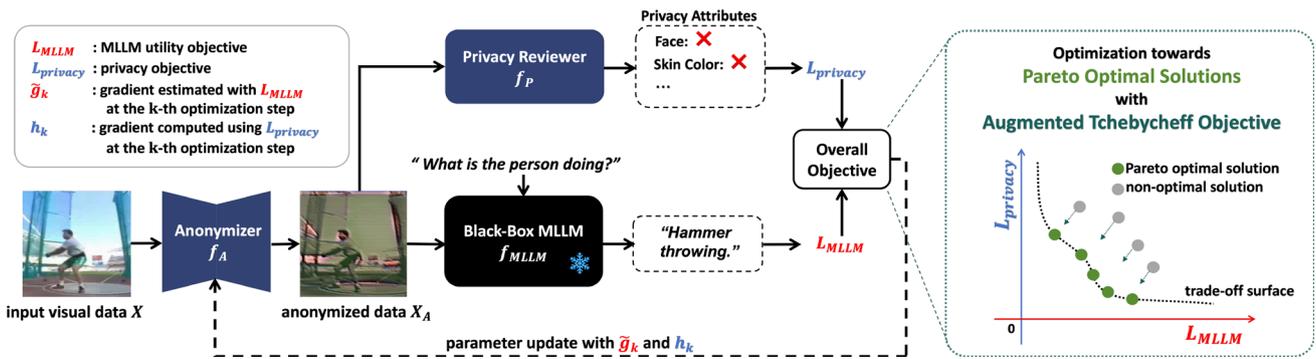

**Fig. 2** Overview of the proposed privacy-preserving framework and optimization pipeline. Given input visual data $X$, the anonymizer $f_A$ generates anonymized data $X_A$. The anonymized data is evaluated by a frozen black-box MLLM $f_{MLLM}$ to obtain the utility objective $L_{MLLM}$ (e.g., with a user instruction such as "What is the person doing?"), and by a privacy reviewer $f_P$ to obtain the privacy objective $L_{Privacy}$. Gradients estimated from $L_{MLLM}$ and computed from $L_{Privacy}$ are used to update the anonymizer. The overall objective is optimized using an augmented Tchebycheff formulation to seek Pareto-optimal trade-offs between privacy and utility

according to the feedback of the black-box function $f$, SPSA estimates the gradient $\hat{g}_k$ at the $k$-th optimization step as:

$$\hat{g}_k = \frac{f(\theta_k + c_k \Delta_k; X) - f(\theta_k - c_k \Delta_k; X)}{2 c_k \Delta_k} \quad (2)$$

where $c_k > 0$ is a positive scalar, and $\Delta_k \in \mathbb{R}^p$ is the random perturbation vector with each of its elements sampled from Bernoulli $\pm 1$ distribution.

### 3.2 Method Overview

In light of the widespread privacy concerns in the usage of MLLM services, we aim to investigate how to protect users' privacy information (e.g., face, skin color, etc.) when using these MLLM services. Notably, we address the practical scenario where the MLLM is a "black box" with no access to its internal information (e.g., model structure and parameters). Our goal is to "erase" the privacy information in the user visual data before uploading it to the MLLM service, with minimum degradation to its performance.

To achieve this, we construct the framework with an anonymizer $f_A$, a privacy reviewer $f_P$, and a frozen black-box MLLM $f_{MLLM}$, as illustrated in Fig. 2. The anonymizer transforms the input visual data $X$ into anonymized data $X_A = f_A(X)$, which is then evaluated to obtain two objectives: (1) the *utility objective* $\mathcal{L}_{MLLM}$ from the black-box MLLM (with smaller values indicating better MLLM performance on the utilization task), and (2) the *privacy objective* $\mathcal{L}_{Privacy}$ from the privacy reviewer (with smaller values indicating less privacy leakage). We learn an optimal anonymizer $f_A$ to achieve the following two criteria: the privacy information in the visual data is maximally removed, and the performance of the MLLM is minimally affected. Thus, the overall goal of our problem is:

$$\min_{\theta} \mathbf{L}(\theta) = \min_{\theta} \left( L_{MLLM}(\theta), L_{Privacy}(\theta) \right), \quad (3)$$

where $\theta \in \mathbb{R}^{p_A}$ is the learnable parameters in the anonymizer $f_A$, and $p_A$ is the number of the learnable parameters. The MLLM is kept frozen. However, achieving this goal can be challenging. First, the anonymizer is required to achieve visual privacy protection while keeping the anonymized data still directly understandable to the frozen MLLM. Yet, generally, stronger privacy-preserving techniques tend to degrade the quality and richness of the original data, which can significantly reduce the MLLM's utility. Moreover, optimizing with black-box MLLM can be difficult to converge, adding to the challenge of achieving our goal. Below, we first discuss how to effectively find a better trade-off between privacy and MLLM's utility, and then investigate how to achieve stable and efficient optimization with the black-box MLLM.

### 3.3 Privacy-Utility Trade-off Learning with Pareto Optimality

In our problem, we aim to optimize the anonymizer $f_A$ with parameters $\theta$ to protect visual privacy while keeping good performance of the black-box MLLM. To effectively address this challenging problem, inspired by Pareto optimality (Miettinen, 1999; Steuer & Choo, 1983), we propose to formulate our goal as finding a *Pareto optimal solution* with the two objectives (i.e., $L_{Privacy}$ and $L_{MLLM}$ in Eq. 3). Under this formulation, a Pareto-optimal solution represents a principled trade-off between privacy preservation and MLLM utility, where neither objective can be improved without sacrificing the other. Naively, to find Pareto optimal solutions, we can formulate the overall objective as a weighted sum of the two objectives:

$$\min_{\theta} L_{ws}(\theta) = \min_{\theta} \lambda_1 L_{MLLM}(\theta) + \lambda_2 L_{Privacy}(\theta) \quad (4)$$

where $\lambda_1 + \lambda_2 = 1$, and $\lambda_1$ and $\lambda_2$ are the weights. While this simplifies the trade-off between these two goals to a single objective, we empirically observe that naively applying the weighted sum objective leads to inferior performance in solving our problem (see results in Tab. 5). This could be due to the unknown characteristics of the black-box MLLM. Specifically, the weighted sum objective is shown to be effective with the assumption that all objectives and the trade-off surface are convex (Geoffrion, 1968). Hence, it struggles to find optimal solutions when the problem space is non-convex, limiting its effectiveness in capturing the optimal trade-off in such a complex scenario as our problem (see Supplementary for more elaboration). As we regard MLLM as a black-box function, there is no guarantee of the convexity of the corresponding objective $L_{MLLM}$ and the trade-off surface. Moreover, in our problem, we obtain two types of gradients for optimizing the two objectives $L_{Privacy}$ and $L_{MLLM}$: (1) for $L_{Privacy}$, we can obtain the computed gradient from the white-box privacy reviewer, but (2) for $L_{MLLM}$, we can only derive the estimated gradient from the black-box MLLM. Thus, the two types of gradients exhibit very different characteristics. Such a gap between the two "unbalanced" gradients further adds uncertainty and challenges to the optimization process of our problem. Therefore, applying the weighted sum objective does not effectively find the optimal solution in our task.

Considering this, in our framework, we aim to specifically enable the overall objective to find a Pareto optimal solution with our complex and possibly non-convex problem. Instead of simply linearly combining the two objectives, we propose to incorporate $L_{MLLM}$ and $L_{Privacy}$ with the help of the Tchebycheff norm:

$$L_{tche}(\theta) = \max_{i} \{\lambda_i (L_{T_i}(\theta) - (z_i^* - \varepsilon))\}, \quad (5)$$

where $L_{T_i} \in \{L_{MLLM}, L_{Privacy}\}$. Minimizing this term means that, at each step, we focus on minimizing the objective that is less satisfied (i.e., more distant from its ideal point). This allows the optimization process to alternately optimize the two objectives, gradually pushing the two objectives to reach their minimum in the learning process. It also breaks the limitation of linearly combining the two objectives through a weighted sum, and enable searching for solutions in non-convex objective spaces.

Moreover, to further ensure the Pareto optimality of our solution, we add an augmentation term to formulate the overall objective as augmented Tchebycheff (AT) objective:

$$\min_{\theta} L_{AT}(\theta) = \min_{\theta} \big( \max_{i} \{\lambda_i (L_{T_i}(\theta) - (z_i^* - \varepsilon))\} \\ + \sigma \sum_{i} \lambda_i L_{T_i}(\theta) \big) \quad (6)$$

where $\sigma$ is a sufficiently small positive scalar. Combining this augmentation term essentially ensures that the solution of Eq. 6 is Pareto optimal as analyzed below:

**Theorem 1** *Let $\theta \in \mathbb{R}^{p_A}$. $\theta$ is Pareto optimal if and only if $\theta$ solves Eq. 6 with $\lambda = [\lambda_1, \lambda_2] \in \mathbb{R}^2$, $\lambda_1, \lambda_2 > 0$.*

Note that while we adopt the normalization convention of the weights ($\lambda_1 + \lambda_2 = 1$), Theorem 1 does not require the weights to be normalized. We provide elaboration of the proof of this theorem in Supplementary. Theorem 1 theoretically motivates the use of the AT objective that optimizing the objective in Eq. 6 leads to a Pareto optimal solution of the problem defined in Eq. 3. In our privacy-preserving problem, a Pareto optimal solution represents the best achievable balance between privacy and utility: we cannot improve privacy protection without degrading MLLM performance, and vice versa. This is crucial because any non-Pareto optimal solution would be wasteful: it would unnecessarily sacrifice privacy or utility when both could be improved simultaneously. By focusing on Pareto optimal solutions, the formulation discourages suboptimal solutions where privacy and utility are unnecessarily sacrificed.

To this end, by designing our overall objective as Eq. 6, we can train the anonymizer to achieve an optimal trade-off between the privacy and the black-box MLLM's utility. Nevertheless, this optimization process can still be challenging due to the difficulty of optimization with the black-box MLLM. Thus, we investigate tackling this challenge in the following.

### 3.4 Critical-History Enhanced Optimization with Black-box MLLM

In this section, we first discuss traditional zeroth-order optimization and analyze its challenges specifically in the context of black-box MLLM. Then, to address these challenges, we propose the critical-history enhanced optimization scheme.

**Zeroth-order optimization.** To achieve optimization with the black-box function, we first turn to the SPSA method (Spall, 1992), as it can achieve gradient estimation with only two forward evaluations of the black-box function at each step. Specifically, in our framework, to optimize the anonymizer $f_A$ with learnable parameters $\theta$ according to the feedback of the black-box MLLM $L_{MLLM}$, SPSA estimates

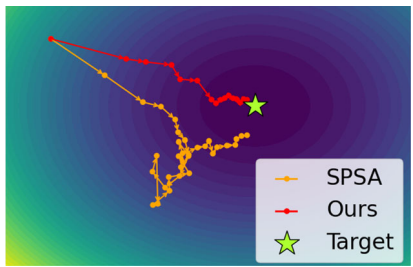

**Fig. 3** Illustration of gradient descent trajectories of SPSA (yellow) and our method (red) on loss landscape. Both optimization start from top-left corner and aim toward same target (green star) (Color figure online)

the gradient $\hat{g}_k$ at the $k$-th optimization step as:

$$\hat{g}_k = \frac{L_{MLLM}(\theta_k + c_k \Delta_k; X) - L_{MLLM}(\theta_k - c_k \Delta_k; X)}{2 c_k \Delta_k} \quad (7)$$

where $c_k > 0$ is a positive scalar and $\Delta_k \in \mathbb{R}^{p_A}$. Essentially, at each optimization step, SPSA randomly perturbs the parameters and evaluates how the random changes impact the performance of the black-box MLLM. Based on the feedback (performance) of the black-box MLLM, it then approximates the optimization direction for adjusting the parameters.

Though SPSA can allow gradient estimation, we empirically find that SPSA optimization can be unstable and slow in convergence, as shown in Fig. 3. This can be because, at each optimization step, SPSA assumes that the target gradient of the estimation is a *truly arbitrary* vector and estimates the gradient by randomly perturbing the parameters. Thus, it inevitably introduces randomness in the search directions and can cause random errors, making the estimation particularly difficult for the high-dimensional problem in our task. Though it is theoretically proved that the random errors can eventually average out after a number of steps and SPSA can converge (Spall, 1992), we find that using SPSA in our problem can even be hard to reach convergence for GPT-4V within two weeks, which can lead to high costs for API calls. This necessitates the design of a more effective scheme to handle optimization with black-box MLLMs in our problem, as follows.

**Critical-history enhanced optimization.** Motivated by the challenges in optimization with the black-box MLLM, we aim to reduce the random errors in the gradient estimation, to enable faster convergence. Particularly, we find that although we cannot access the internal information of the black-box MLLM, the target gradient of the estimation is not necessarily a *totally arbitrary* vector. In fact, at each optimization step, the gradient estimation can be seen as an exploration of the landscape of the black-box MLLM objective. Thus, throughout the optimization process, we can actually accumulate knowledge about the black-box MLLM from previous explorations. This accumulated knowledge can provide valuable prior information, which can be used to mitigate the randomness in estimating the target gradient.

Thus, in our framework, we take advantage of the history gradients in past optimization steps and propose *critical-history enhanced optimization*. In this scheme, we improve the gradient estimation by: 1) specifically collecting history gradients that contain more critical information, and 2) using the collected critical history gradients as prior information to correct and adjust optimization at the current step.

Collection of critical gradients. Due to the randomness in gradient estimation, the estimated gradients in previous steps may contain varying level of useful information, i.e., the gradients in some steps may contribute more to the optimization, while in some steps, the gradients can be noisier and less useful. Thus, we aim to examine the estimated gradient and collect the most informative ones.

To achieve this, as the gradient's contribution to the optimization is dependent on the gradient norm (Paul et al., 2021; Mcrae et al., 2022), we leverage the L2 norm as the indicator to determine the importance of the estimated gradients. Also, to make sure that the selected gradients are effective, we apply a decay factor to reduce their impact over time. More concretely, we keep a gradient collection set $C_g$ with size $m$. At each step $k$, we obtain the estimated gradient $\hat{g}_k$ via Eq. 7. Then, we compute the importance score for the current gradient $\hat{g}_k$, and update the importance scores for each of the previous gradients $\hat{g}_{t_i}$ in $C_g$ with a decay factor $d \in (0, 1]$:

$$s_k = \|\hat{g}_k\|_2, \quad s_{t_i} = d^{k-t_i} \|\hat{g}_{t_i}\|_2. \quad (8)$$

In this way, we obtain importance scores of all previous gradients in $C_g$: $S = [s_{t_1}, s_{t_2}, ..., s_{t_m}]$. If the importance score of the current gradient $s_k$ is greater than the smallest importance score in $S$, i.e., $s_k > s_{t_n}$, $s_{t_n} = \min S$, we remove the history gradient $\hat{g}_{t_n}$ and add the current gradient $\hat{g}_k$ to the collection set $C_g$. By doing so, we keep a set of critical gradients that contain important and timely information.

Optimization with history gradient. With the collected critical history gradients as prior information, we propose to guide the optimization with the black-box MLLM by improving the *gradient direction* and adjusting the *step size*. In particular, at step $k$, with the collected $m$ critical history gradients in $C_g$, we can improve gradient directions and compute the enhanced estimated gradient $\tilde{g}_k$ as:

$$\tilde{g}_k = \gamma \hat{g}_k + \frac{1-\gamma}{m} \sum C_g, \quad \tilde{g}_1 = \hat{g}_1, \quad (9)$$

where $\gamma \in (0, 1)$ controls the impact of the history gradients. This can help to obtain more accurate optimization directions: the enhanced estimated gradient $\tilde{g}_k$ is estimated with the information in $m$ critical history gradients in addition to the current step, rather than relying solely on the current step, leading to a more stable optimization.

Moreover, besides improving the gradient directions, we also adaptively adjust the *step size* according to the optimization status. Specifically, we propose to assess the optimization status at each step and adjust the step size accordingly. To evaluate this status, we compare the directions between the estimated gradient $\hat{g}_k$ and the enhanced estimated gradient $\tilde{g}_k$. Denoting the angle between $\hat{g}_k$ and $\tilde{g}_k$ as $\varphi_k$, we can compute the status indicator $I_k$ at the $k$-th step as:

$$I_k = \frac{b}{\pi} \arctan(2\pi \cos \varphi_k), \qquad (10)$$

where $b \in (0, 2)$ is the factor controlling the magnitude of $I_k$, and $I_k$ is within the range of $(-\frac{b}{2}, \frac{b}{2})$. Larger $I_k$ indicates that $\hat{g}_k$ and $\tilde{g}_k$ are more aligned, i.e., the current gradient estimation is more consistent with the critical gradients in previous steps. Conversely, smaller $I_k$ suggests that the directions of $\hat{g}_k$ and $\tilde{g}_k$ diverge, and the optimization direction in the current step may be noisier and deviate from the correct path. In this way, we can use $I_k$ to reflect the current optimization status. Then, we adaptively adjust the original step size $a_k$ with the status indicator $I_k$ as $a'_k = (1 + I_k)a_k$. By doing so, we can reduce the impact of random errors in noisy estimations, which can further stabilize and improve the overall optimization.

Overall, in the critical-history enhanced optimization, we collect critical history gradients from previous optimization steps, and use them as prior information to improve gradient directions and adjust step sizes. These designs all facilitate to reduce the random errors in the gradient estimation, and achieve a more stable and efficient optimization with the black-box MLLM.

### 3.5 Overall Training and Testing

During training, we update the parameters $\theta$ in the anonymizer $f_A$ such that we can achieve an optimal trade-off between the utilization performance of MLLM and privacy preservation. Specifically, we formulate this optimization problem as seeking a Pareto optimal solution using the overall objective $L_{AT}$ in Eq. 6. In each iteration $k$, we obtain the gradient $\hat{\nabla} L_{AT}$ with objective $L_{AT}$:

$$\hat{\nabla} L_{AT}(\theta_k) = \begin{cases} \lambda_1(1 + I_k)\tilde{g}_k + \sigma g, & \text{if } i_{max} = 1 \\ \lambda_2 h_k + \sigma g, & \text{if } i_{max} = 2 \end{cases} \qquad (11)$$

where $i_{max} = \arg\max_i \{\lambda_i (L_{T_i}(\theta) - (z_i^* - \varepsilon))\}$, $g = \lambda_1(1 + I_k)\tilde{g}_k + \lambda_2 h_k$, and $h_k$ is the gradient computed using the privacy preservation objective $L_{Privacy}$. Then, we update the parameter $\theta$ accordingly:

$$\theta_{k+1} = \theta_k - a_k \hat{\nabla} L_{AT}(\theta_k). \qquad (12)$$

We show the detailed algorithm to obtain the estimated gradient $\tilde{g}_k$ and status indicator $I_k$ in Supplementary. During testing, we first process the input visual data using the anonymizer $f_A$ and generate the anonymized data, which is then passed to the MLLM to perform the utilization task. We also evaluate the performance of privacy protection using the privacy reviewer.

## 4 Experiments

To evaluate our proposed framework, we first conduct experiments on the well-established privacy-preserving action recognition benchmarks (Wu et al., 2020; Dave et al., 2022). To further evaluate our framework, we also conduct experiments with the common MLLM task visual question answering (VQA) (Antol et al., 2015). In our experiments, we consider the MLLM as a black box, with no information about its structure and parameters.

### 4.1 Datasets and Evaluation Metrics

First, following the well-established privacy-preserving action recognition benchmark (Dave et al., 2022; Wu et al., 2020; Peng et al., 2023), we evaluate our framework with two settings. (1) We conduct experiments with *same-dataset evaluation* (HMDB51-VISPR) (Wu et al., 2020; Dave et al., 2022), where Kuehne et al. (2011) is a common video action recognition datasets, and Orekondy et al. (2017) is a privacy attribute dataset. During training, following (Wu et al., 2020; Dave et al., 2022), the utility objective and privacy objective are obtained on HMDB51 and VISPR training sets respectively, and during testing, the action recognition performance and privacy performance are evaluated using PA-HMDB (Wu et al., 2020), which contains both action class and privacy attribute annotations. (2) We also follow the *cross-dataset training and evaluation* (UCF101-VISPR) (Dave et al., 2022), where Soomro et al. (2012) is a common action recognition dataset. During training, following (Wu et al., 2020; Dave et al., 2022), the objectives of utility and privacy are obtained on UCF101 and VISPR training sets respectively, and during testing, the action recognition performance is evaluated using UCF101 test set, and the privacy performance is evaluated using VISPR test set (Wu et al., 2020; Dave et al., 2022).

Moreover, we also evaluate our framework with the VQA dataset OK-VQA (Marino et al., 2019), which has been commonly used to evaluate MLLMs (Huang & Zhang, 2024; Liu et al., 2023a; Chen et al., 2023). We also follow the two eval-uation settings (i.e., same-dataset and cross-dataset) (Dave et al., 2022) to train and evaluate our framework on VQA task. For all experiments, we follow the definition of privacy (i.e., privacy attributes) in previous works (Dave et al., 2022; Peng et al., 2023) (see more details for the benchmarks in Supplementary).

Following (Wu et al., 2018, 2020; Peng et al., 2023), we evaluate privacy preservation performance using privacy attribute classifier (i.e., the privacy reviewer) with class-wise mean average precision (cMAP). We report the Top-1 accuracy for action recognition following (Wu et al., 2018, 2020; Peng et al., 2023), and evaluate VQA performance using VQA accuracy score following (Antol et al., 2015; Marino et al., 2019) (see Supplementary for more details about the metrics).

### 4.2 Implementation Details

We construct the anonymizer $f_A$ based on Ronneberger et al. (2015), which has shown its effectiveness in previous privacy-preserving works (Wu et al., 2020; Dave et al., 2022; Peng et al., 2023), and follow (Wu et al., 2020; Dave et al., 2022) to adopt ResNet-50 (He et al., 2016) as the privacy reviewer $f_P$. We follow the training scheme in Wu et al. (2020, 2018); Peng et al. (2023) for $f_P$. To evaluate our framework, we conduct the following two groups of experiments: (1) As it can be costly and inconvenient to conduct experiments on commercial cloud MLLMs such as GPT-4V, we first adopt popular MLLMs Video-LLaVA Lin et al. (2023) (for video action recognition) and Liu et al. (2023a) (for image VQA). We deploy Video-LLaVA and LLaVA locally and regard them as black-box models to conduct comprehensive experiments. (2) Then, to further explore the real-world scenarios, we also evaluate our framework with commercial cloud MLLM service GPT-4V (Achiam et al., 2023). For all experiments, we follow standard SPSA (Spall, 1992) to assign $a_k$ and $c_k$. Our main experiments are conducted on an RTX 3090 GPU. We show the details regarding $L_{MLLM}$, $L_{Privacy}$, $f_A$, $f_P$, and more details about the implementation and hyperparameters in Supplementary.

### 4.3 Experiments with Action Recognition

Following (Wu et al., 2020; Dave et al., 2022), we evaluate our framework and compare it with *Downsample* and *Obfuscation-based* methods (Dave et al., 2022) that use heuristic rules to protect privacy information, and *previous privacy-preserving methods* (Wu et al., 2020; Dave et al., 2022; Kumawat & Nagahara, 2022; Peng et al., 2023; Li et al., 2023; Ilic et al., 2024). Specifically, to enable (Wu et al., 2020; Dave et al., 2022; Kumawat & Nagahara, 2022; Peng et al., 2023; Li et al., 2023) to optimize with the black-box MLLM, we follow (Liu et al., 2016) and consider the action recognition model in the each corresponding framework as the surrogate model for the black-box MLLM. The surrogate model is trained on action recognition task with cross-entropy loss. More discussion about the methods and implementation details are in Supplementary. We report the results using Video-LLaVA and GPT-4V in Tab. 1 and Tab. 2 respectively. We observe that the compared methods typically fail to achieve a robust privacy-utility balance in the black-box MLLM setting. Specifically, methods such as obfuscation-based baselines tend to result in substantial drops in MLLM performance. This can be because these methods suppress privacy attributes by degrading visual content, which can also remove task-relevant information and thus degrade MLLM performance. On the other hand, methods that require white-box access to the utility model often struggle to achieve good privacy-utility trade-off, which is possibly because these methods fail to learn meaningful anonymization strategy tailored to the black-box MLLM, as they do not directly support optimization with black-box MLLM. In contrast, our method achieves top performance on the utilization task while having better privacy preservation performance. Overall, our proposed framework achieves the best trade-off between privacy and utility in experiments with both Video-LLaVA and GPT-4V.

### 4.4 Experiments with Visual Question Answering

To better evaluate our framework, we also conduct experiments with the VQA task, which is commonly used to evaluate the performance of MLLMs (Liu et al., 2023a; Chen et al., 2023). Specifically, we adopt the subcategories of OK-VQA with images that are most critical to privacy leakage, i.e., *People and Everyday Life* and *Sports and Recreation*. We annotate the VQA testing images with the same privacy attributes following (Dave et al., 2022) and evaluate our framework following cross-dataset training and same-dataset evaluation setting (Dave et al., 2022; Wu et al., 2020). Following (Dave et al., 2022; Wu et al., 2020), the objectives of utility and privacy are obtained with OK-VQA and VISPR training samples respectively, and the performance of VQA and privacy is evaluated using the annotated OK-VQA testing samples. Also, following the cross-dataset training and evaluation setting (Dave et al., 2022; Wu et al., 2020), we evaluate our framework using VISPR testing samples. We compare our framework with methods introduced in Sec. 4.3. Specifically, for Wu et al. (2020); Dave et al. (2022); Kumawat and Nagahara (2022); Peng et al. (2023); Li et al. (2023), we adopt VQA model (Gardères et al., 2020) pre-trained on OK-VQA as the surrogate model. More details about the annotation and comparison methods are in Supplementary. We report our results using LLaVA and GPT-4V in Tab. 3 and Tab. 4 respectively. As shown, our method can achieve the best trade-off between the VQA accuracy and privacy protection, showing its effectiveness.

**Table 1** Results on action recognition using Video-LLaVA. Note that, Top-1 action recognition results for "Raw data" are obtained by feeding the original testing data (video frames) to the publicly available large model Video-LLaVA. Other Top-1 action recognition results are obtained by feeding the anonymized testing data (produced using the compared methods and our method) to the same fixed Video-LLaVA. As shown, our method protects privacy while least affecting the large model's performance

| Method | HMDB51-VISPR | | UCF101-VISPR | |
| --- | --- | --- | --- | --- |
| | Action Top-1 ($\uparrow$) | Privacy cMAP ($\downarrow$) | Action Top-1 ($\uparrow$) | Privacy cMAP ($\downarrow$) |
| Raw data | 50.3 | 70.1 | 55.4 | 64.4 |
| Downsample Dave et al. (2022) | 42.1 | 61.2 | 43.1 | 57.2 |
| Obf-Blackening Dave et al. (2022) | 26.6 | 63.8 | 40.1 | 56.4 |
| Obf-StrongBlur Dave et al. (2022) | 27.2 | 64.4 | 43.6 | 55.9 |
| Obf-WeakBlur Dave et al. (2022) | 33.2 | 69.4 | 46.0 | 63.5 |
| Surrogate model of VITA Wu et al. (2020) | 31.6 | 63.5 | 34.2 | 57.1 |
| Surrogate model of SPAct Dave et al. (2022) | 32.4 | 62.9 | 36.8 | 56.7 |
| Surrogate model of BDQ Kumawat and Nagahara (2022) | 33.1 | 63.4 | 35.6 | 56.4 |
| Surrogate model of MPPAR Peng et al. (2023) | 32.6 | 63.2 | 36.9 | 57.4 |
| Surrogate model of STPrivacy Li et al. (2023) | 32.3 | 62.5 | 37.4 | 57.1 |
| SelectivePrivacy Ilic et al. (2024) | 35.2 | 63.5 | 39.8 | 59.1 |
| Ours | **47.6** | **59.8** | **48.4** | **54.6** |

**Table 2** Results on action recognition using GPT-4V

| Method | HMDB51-VISPR | | UCF101-VISPR | |
| --- | --- | --- | --- | --- |
| | Action Top-1 ($\uparrow$) | Privacy cMAP ($\downarrow$) | Action Top-1 ($\uparrow$) | Privacy cMAP ($\downarrow$) |
| Raw data | 60.6 | 70.1 | 70.4 | 64.4 |
| Downsample Dave et al. (2022) | 51.7 | 61.2 | 61.5 | 57.2 |
| Obf-Blackening Dave et al. (2022) | 47.6 | 63.8 | 62.4 | 56.4 |
| Obf-StrongBlur Dave et al. (2022) | 48.1 | 64.4 | 61.7 | 55.9 |
| Obf-WeakBlur Dave et al. (2022) | 50.4 | 69.4 | 63.1 | 63.5 |
| Surrogate model of VITA Wu et al. (2020) | 48.3 | 63.2 | 58.9 | 58.5 |
| Surrogate model of SPAct Dave et al. (2022) | 47.8 | 62.1 | 59.1 | 57.9 |
| Surrogate model of BDQ Kumawat and Nagahara (2022) | 47.5 | 62.4 | 58.7 | 58.1 |
| Surrogate model of MPPAR Peng et al. (2023) | 48.6 | 62.2 | 59.2 | 57.8 |
| Surrogate model of STPrivacy Li et al. (2023) | 48.8 | 61.7 | 59.8 | 57.4 |
| SelectivePrivacy Ilic et al. (2024) | 48.5 | 62.8 | 61.2 | 59.1 |
| Ours | **54.3** | **57.3** | **67.1** | **54.9** |

### 4.5 Ablation Studies and Visualizations

We evaluate the designs of our method on UCF101-VISPR following (Dave et al., 2022; Wu et al., 2018). **More ablation studies, visualizations, and further analysis are provided in Supplementary.**

**Impact of the augmented Tchebycheff objective $L_{AT}$.** In our framework, we design the overall objective as augmented Tchebycheff to achieve optimal trade-off between privacy and utility. We evaluate this design and compare with the following variants: **w/ weighted sum** variante that only uses Eq. 4 as the objective, and **w/ Tchebycheff norm** variant that only uses Eq. 5 as the objective. We report the best result for the variants. As shown in Tab. 5, using weighted sum objective is not effective to solve our problem, and using augmented Tchebycheff objective can achieve the best trade-off, showing the efficacy of our design.

**Comparisons of zeroth-order optimization methods.** We propose the critical-history enhanced optimization to improve the optimization with the black-box MLLM. We compare this design with other zeroth-order optimization methods. We first compare with the original SPSA and its recent variant SPSA-GC (Oh et al., 2023). As shown in Fig.

**Table 3** Results on VQA using LLaVA

| Method | OK-VQA VQA Accuracy (↑) | OK-VQA Privacy cMAP (↓) | VISPR Privacy cMAP (↓) |
|---|---|---|---|
| Raw data | 60.1 | 59.3 | 64.4 |
| Downsample Dave et al. (2022) | 55.7 | 52.7 | 57.2 |
| Obf-Blackening Dave et al. (2022) | 48.4 | 51.3 | 56.4 |
| Obf-StrongBlur Dave et al. (2022) | 54.4 | 50.9 | 55.9 |
| Obf-WeakBlur Dave et al. (2022) | 55.9 | 57.5 | 63.5 |
| Surrogate model of VITA Wu et al. (2020) | 40.2 | 49.3 | 54.3 |
| Surrogate model of SPAct Dave et al. (2022) | 41.6 | 48.6 | 52.9 |
| Surrogate model of MPPAR Peng et al. (2023) | 43.1 | 48.2 | 52.1 |
| SelectivePrivacy Ilic et al. (2024) | 45.1 | 51.5 | 54.3 |
| Ours | **57.6** | **44.2** | **48.8** |

**Table 4** Results on VQA task using GPT-4V

| Method | OK-VQA VQA Accuracy (↑) | OK-VQA Privacy cMAP (↓) | VISPR Privacy cMAP (↓) |
|---|---|---|---|
| Raw data | 63.6 | 59.3 | 64.4 |
| Downsample Dave et al. (2022) | 56.2 | 52.7 | 57.2 |
| Obf-Blackening Dave et al. (2022) | 49.1 | 51.3 | 56.4 |
| Obf-StrongBlur Dave et al. (2022) | 55.3 | 50.9 | 55.9 |
| Obf-WeakBlur Dave et al. (2022) | 56.1 | 57.5 | 63.5 |
| Surrogate model of VITA Wu et al. (2020) | 55.4 | 48.4 | 50.2 |
| Surrogate model of SPAct Dave et al. (2022) | 56.5 | 46.6 | 49.4 |
| Surrogate model of MPPAR Peng et al. (2023) | 55.6 | 46.8 | 49.7 |
| SelectivePrivacy Ilic et al. (2024) | 55.9 | 51.5 | 54.3 |
| Ours | **59.2** | **44.2** | **47.9** |

**Table 5** Impact of augmented Tchebycheff

| Method | UCF101-VISPR | |
|---|---|---|
| | Action (Top-1 ↑) | Privacy (cMAP ↓) |
| w/ weighted sum (Eq. 4) | 32.9 | 58.4 |
| w/ Tchebycheff norm (Eq. 5) | 42.6 | 57.3 |
| w/ augmented Tchebycheff (Eq. 6) | 48.4 | 54.6 |

**Table 6** Comparisons of zeroth-order optimization methods on UCF101-VISPR using Video-LLaVA as the black-box MLLM. We report action recognition performance, privacy performance, and the average number of MLLM calls per sample during training

| Method | UCF101-VISPR | | No. of MLLM calls (per sample) |
|---|---|---|---|
| | Action (Top-1 ↑) | Privacy (cMAP ↓) | |
| ES Hansen et al. (2003) | 30.2 | 59.2 | 161 |
| RL Watkins and Dayan (1992) | 32.6 | 61.3 | 147 |
| SPSA Spall (1992) | 35.7 | 57.5 | 136 |
| SPSA-GC Oh et al. (2023) | 37.9 | 57.1 | 80 |
| Ours | 48.4 | 54.6 | 24 |

Table 7 Impact of critical-history enhanced scheme

| Method | UCF101-VISPR | |
|---|---|---|
| | Action (Top-1 ↑) | Privacy (cMAP ↓) |
| w/o history gradient | 35.7 | 57.5 |
| w/o critical gradient collection | 44.2 | 56.7 |
| w/o gradient direction improvement | 43.4 | 57.2 |
| w/o step size adjustment | 43.9 | 56.9 |
| Ours | 48.4 | 54.6 |

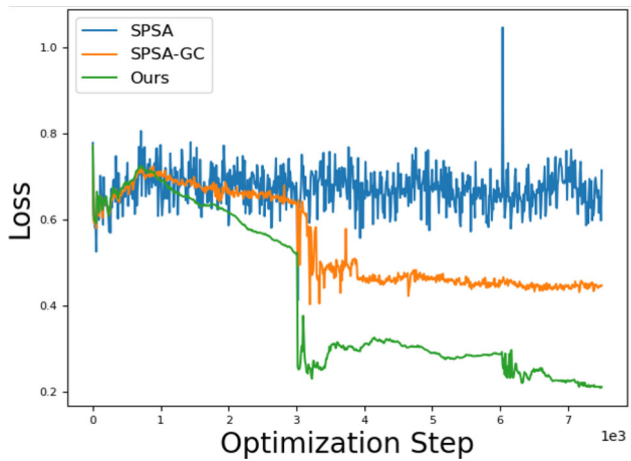

**Fig. 4** Comparisons of loss value changes over optimization steps

4, our critical-history enhanced optimization achieves much faster and more stable optimization, which is more suitable for our problem. We also compare other common methods for optimization with the black-box function, including evolution strategies (*ES*) Hansen et al. (2003), and reinforcement learning (*RL*) Watkins and Dayan (1992), and report the number of MLLM calls required per sample during training. As shown in Tab. 6, our proposed method achieves the best results while significantly reducing the number of API calls.

**Impact of the critical-history enhanced optimization**. To evaluate the design of the critical-history enhanced optimization, we conduct experiments with the following variants: (1) **w/o history gradient**: we discard history gradients and do not adjust gradient direction and step size. (2) **w/o critical gradient collection**: we use the latest $m$ history gradients without selection to adjust gradient directions and step sizes. (3) **w/o gradient direction adjustment**: we do not adjust gradient directions and only adjust step sizes using the critical gradients. (4) **w/o step size adjustment**: we do not adjust step sizes and only adjust gradient directions using the critical gradients. As shown in Tab. 7, our proposed method show the best performances on both action recognition and privacy protection, demonstrating the effectiveness of our design.

**Qualitative results**. We show visualization results of the video frames before and after our anonymizer $f_A$ for the action recognition task in Fig. 5. We also provide the visualization results for the VQA task in Fig. 6. We show more anonymized images generated by our method on VISPR testing set in Fig. 7, which contains annotations of privacy information such as nudity and credit card. As shown, our anonymizer can effectively remove the privacy information within the visual data, while the MLLM can still give correct predictions with the anonymized data.

**Discussion on scenarios with constrained budget on API usage.** Our framework is designed with API efficiency in mind from the outset. In particular, the proposed critical-history enhanced optimization substantially reduces the number of MLLM calls required for convergence, while consistently achieving better privacy–utility trade-offs than existing zeroth-order alternatives (Tab. 6). Beyond this inherent efficiency, the framework also naturally supports additional strategies to further adapt to budget-constrained scenarios. Specifically, in low-budget scenario, increasing the size of the history gradient collection set $m$ can allow greater reuse of informative past gradients, thereby reducing the need for new MLLM queries. Second, we can employ adaptive scheduling to allocate the API budget non-uniformly across samples, prioritizing those with higher uncertainty or slower convergence rates, thereby avoiding unnecessary queries on already well-optimized samples. Finally, early stopping criteria based on the stabilization of the optimization trajectory can be applied to terminate the training once marginal performance gains diminish. With these strategies, our method can further reduce API calls in constrained budget scenario, while still achieving robust performance.

## 5 Conclusion

In this paper, we are the first to investigate how to protect visual privacy when using black-box MLLMs. To tackle this problem, we propose a novel privacy-preserving framework for black-box MLLM. Specifically, to effectively find a better trade-off with the black-box MLLM, we propose to design the optimization objective with Pareto optimality. Meanwhile, in this process, we propose to ease the challenges

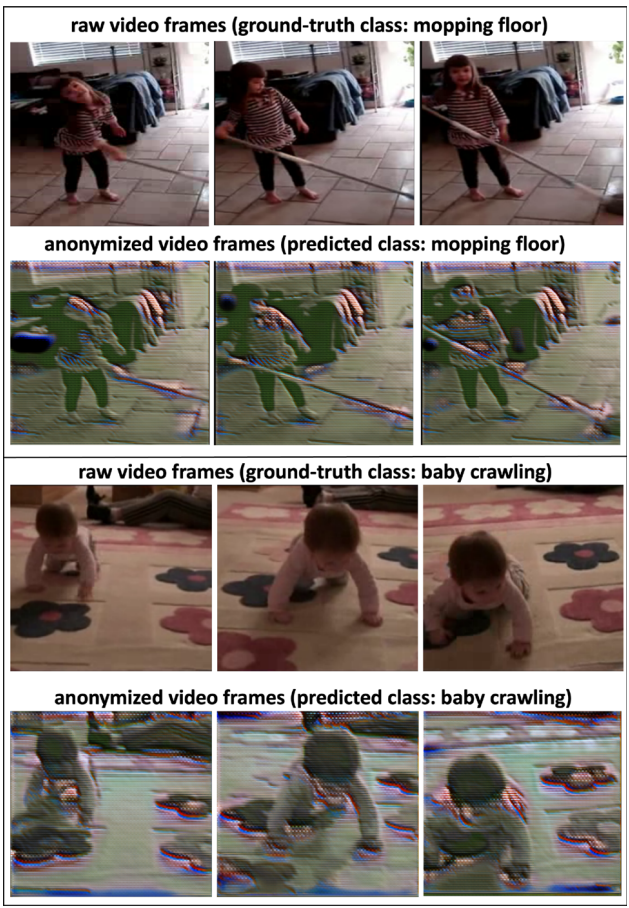

**Fig. 5** Visualization results of action video frames. In each row, we show the raw action video frames and anonymized video frames generated by our anonymizer, along with the ground-truth and predicted action class. As shown, the anonymizer in our framework can remove privacy information such as face and skin color, while the MLLM can still give correct predictions

of optimization with black-box MLLM and design the critical history enhanced optimization scheme. Our experiments show the effectiveness of our framework with both image and video utilization tasks.

## 6 Future Work

While this work takes an initial step toward privacy preservation for black-box MLLMs, it also has several limitations that point to promising directions for future research. First, our current evaluation of privacy preservation primarily relies on cMAP, which is a commonly used metric for quantifying privacy leakage. While cMAP provides a useful and standardized measure, it may not fully capture the fine-grained semantics or varying severities of privacy leakage in real-world scenarios. More fine-grained and semantically meaningful privacy metrics could be explored to better cap-ture different types and severities of privacy leakage. Second, improving the generalization ability of the anonymizer to new domains, unseen privacy attributes, and novel utilization tasks is an important direction, particularly for real-world deployment. Third, though our method substantially reduces the API calls during training, more efficient training strategies under strict API budget constraints are worth further investigation. We hope this work can stimulate broader interest in developing advanced privacy-preserving machine learning and make MLLM services safer and more trustworthy for users.

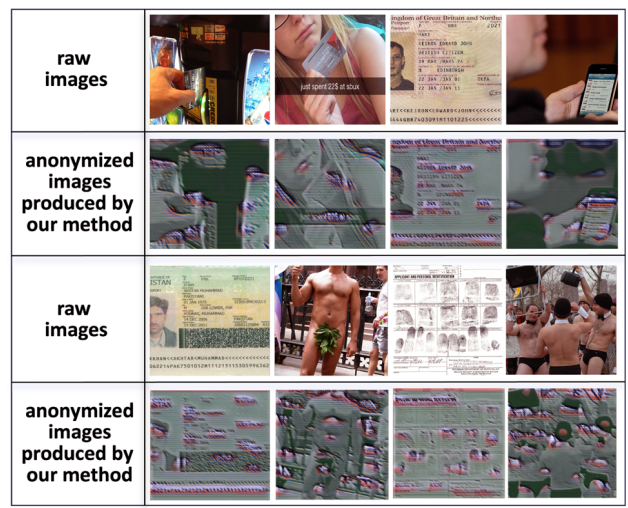

**Fig. 6** Visualization results of VQA images. For each example, we show the raw image and the anonymized image generated by our anonymizer. We pass the question with the anonymized image to the MLLM. The question, MLLM answer, and the annotated answer provided by the dataset are listed for each example. An MLLM answer is considered correct if it can match to the annotated answers following the evaluation rule in Antol et al. (2015); Chen et al. (2023). The MLLM answers in the above three examples are all correct

**Fig. 7** Visualizations of our generated anonymized images on VISPR testing set. The VISPR dataset includes annotations for various privacy attributes, such as faces, nudity, and credit card details. As shown in the figure, our method effectively removes the targeted privacy information

**Supplementary Information** The online version contains supplementary material available at https://doi.org/10.1007/s11263-026-02761-y.

**Data Availability** The datasets that used in the experiments in this paper are available in reference number Orekondy et al. (2017); Kuehne et al. (2011); Soomro et al. (2012); Wu et al. (2020); Marino et al. (2019). These datasets are available at: https://tribhuvanesh.github.io/vpa/, https://serre-lab.clps.brown.edu/resource/hmdb-a-large-human-motion-database/, https://www.crcv.ucf.edu/data/UCF101.php, https://github.com/VITA-Group/PA-HMDB51, https://okvqa.allenai.org. Our privacy attribute annotation for VQA task is available at https://drive.google.com/file/d/1gshmvM2Y9zC9mfN_Rvo2wOlhKRpIDNw5/view?usp=share_link.